\documentclass{article}
\usepackage{spconf,amsmath,graphicx}
\usepackage{amsfonts}
\usepackage{seqsplit}

\title{Enhancing Zero-shot Personalized Image Aesthetics Assessment with Profile-aware Multimodal LLM}

\name{
Chun Wang$^{\star}$\thanks{Contact: Chun Wang, lukewang25@live.cn} \qquad
Chenfeng Wei$^{\dagger}$ \qquad
Chenyang Liu$^{\star}$ \qquad
Weihong Deng$^{\ddagger }$
}

\address{
$^{\star}$ Mashang Consumer Finance Co., Ltd., Chongqing, China \\
$^{\dagger}$ Xi'an Jiaotong-Liverpool University, Suzhou, China \\
$^{\ddagger}$ Beijing University of Posts and Telecommunications, Beijing, China 
}

\begin{document}
%
\maketitle

\begin{abstract}
Personalized image aesthetics assessment (PIAA) aims to predict an individual user's subjective rating of an image, which requires modeling user-specific aesthetic preferences. Existing methods rely on historical user ratings for this modeling and therefore struggle when such data are unavailable. We address this zero-shot setting by using user profiles as contextual signals for personalization and adopting a profile-based personalization paradigm. We introduce P-MLLM, a profile-aware multimodal LLM that augments a frozen LLM with selective fusion modules for controlled visual integration. These modules selectively integrate visual information into the model's evolving hidden states during profile-conditioned reasoning, allowing visual information to be incorporated in a profile-aware manner. Experiments on recent PIAA benchmarks show that P-MLLM achieves competitive zero-shot performance and remains effective even with coarse profile information, highlighting the potential of profile-based personalization for zero-shot PIAA.
\end{abstract}

\section{Introduction}
\label{sec:intro}

\begin{figure}[t]
\centering
\begin{minipage}[b]{1.0\linewidth}
  \centering
  \centerline{\includegraphics[width=1.0\linewidth]{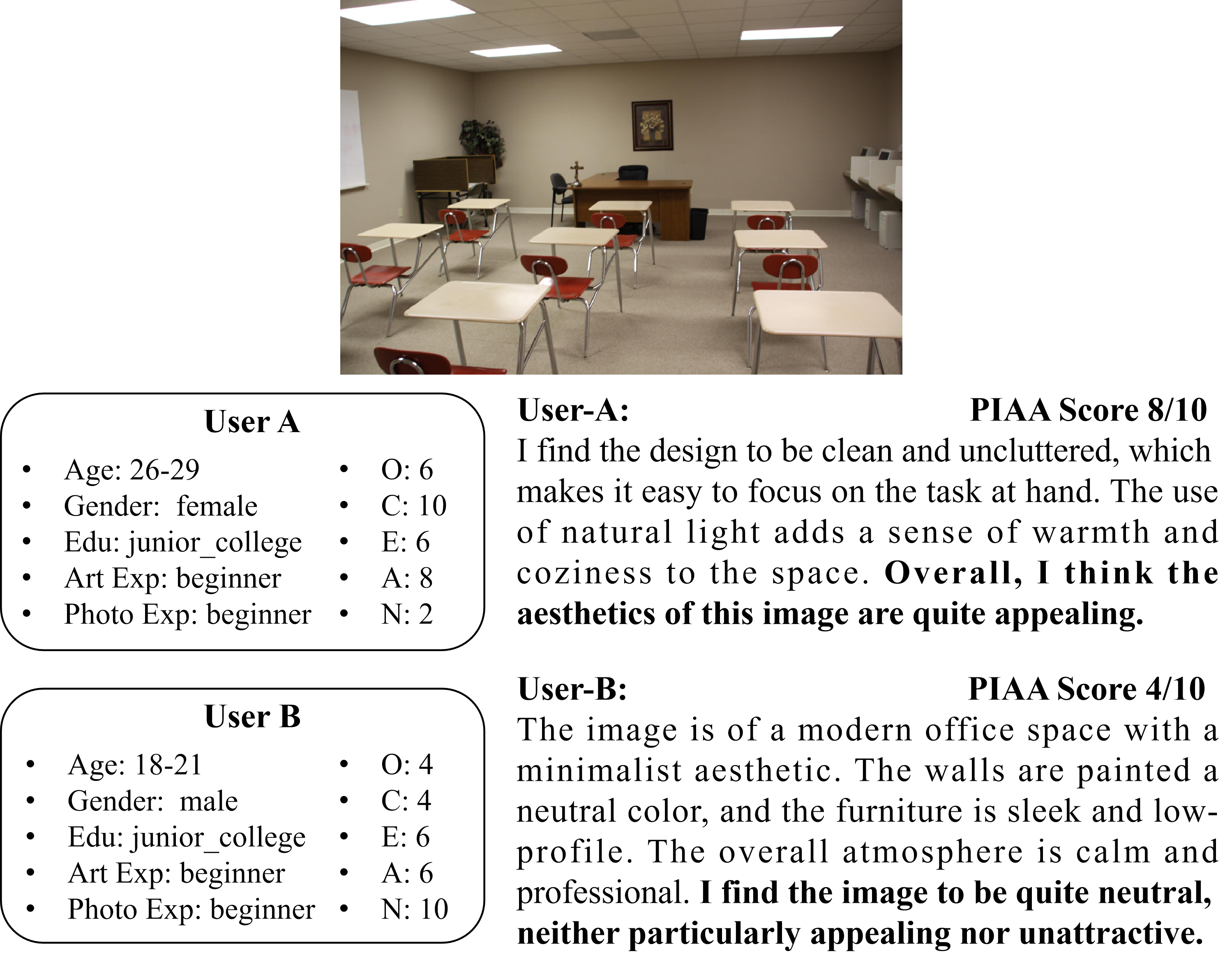}}
\end{minipage}
\caption{An illustration of the PIAA task.}
\label{fig:piaa}
\vspace{-1em}
\end{figure}

Personalized image aesthetics assessment (PIAA) aims to predict user-specific aesthetic judgments for images and is increasingly relevant to human-centric applications such as personal photo management and personalized content recommendation \cite{PARA}. As illustrated in Fig.~\ref{fig:piaa}, users with different backgrounds or personality traits may assign notably different ratings to the same image, reflecting the inherently subjective and user-dependent nature of aesthetic perception.

The core challenge of PIAA lies in modeling user-specific aesthetic preferences to enable personalized predictions. Existing approaches typically rely on each user's historical aesthetic-rating data to achieve personalized modeling. Because such per-user data are often scarce, most methods \cite{PARA}\cite{Ren_2017_ICCV} first train a generic image aesthetics assessment (GIAA) model to capture population-level preferences and then adapt it to individual users to improve data efficiency. However, this reliance on user-specific aesthetic ratings limits the broader applicability of these methods, as such PIAA-specific user behavior data are rarely available for most users, leading to a zero-shot scenario\footnote{The term *zero-shot* means no target‑user task‑specific data; cold‑start is equally applicable.} where no target-user ratings exist to guide personalization. This motivates exploring alternative forms of user information and new paradigms for personalization.

A natural alternative is user profiles that capture information such as a user's background or personality traits. These profiles can be collected through self-descriptions or inferred from behavior in other domains, making them more accessible than PIAA-specific aesthetic ratings\footnote{The profile can be viewed as a medium that enables cross‑task personalization.}. Recent advances in large language models (LLMs) show that profile-based prompting can produce distinct responses conditioned on profile semantics, suggesting that profiles can serve as a useful signal for personalization \cite{liu2025surveypersonalizedlargelanguage}. However, applying profile-based prompting to PIAA is nontrivial: although LLMs can condition on profile semantics in textual tasks, PIAA is inherently multimodal and requires aligning profile information with visual inputs to support profile-conditioned reasoning over images, motivating a mechanism for profile-conditioned visual information fusion that enables personalized image aesthetics prediction.

In this paper, we study a profile-based personalization paradigm for PIAA and introduce P-MLLM, a profile-aware multimodal LLM designed for the zero-shot setting where no aesthetic ratings from the target user are available. Our approach leverages the inherent profile-conditioned behavior of foundation models and adapts it to the task of subjective aesthetic prediction through a selective fusion framework. P-MLLM is constructed by augmenting a frozen LLM with lightweight selective fusion modules that integrate visual information into the model’s evolving hidden states during profile-conditioned reasoning. This design enables the model to produce profile-conditioned variations in its image-related outputs and support user-specific aesthetic predictions. Experiments on two recent PIAA datasets demonstrate that P-MLLM achieves competitive zero-shot performance and remains effective even when provided only coarse profile information. The main contributions are:

\begin{itemize}
  \item We study a profile-based personalization paradigm for zero-shot PIAA, where user profiles serve as an alternative source of personalization signals and enable prediction for users without historical aesthetic ratings.
  \item We propose P-MLLM, a profile-aware multimodal LLM that performs profile-conditioned selective fusion to support personalized image aesthetics prediction.

\end{itemize}

\section{Related Works}
\label{sec:related}


\subsection{Personalized Image Aesthetics Assessment}
\label{ssec:piaa}
To address the data scarcity challenge, early PIAA approaches often train a general GIAA model first and then fine-tune it into user-specific PIAA models using limited per-user data \cite{Ren_2017_ICCV}, while meta-learning methods improve adaptation efficiency with only a few samples \cite{8803119}. Subsequent works further enhance personalization by modeling user-image interactions \cite{math10224181}, disentangling shared and user-specific cues through contrastive learning \cite{10168279}, or incorporating personality attributes via multi-task learning \cite{8970458}. Although some studies \cite{PARA}\cite{8970458} introduce additional user information as auxiliary signals, existing methods fundamentally depend on user-specific aesthetic ratings to capture individual aesthetic preferences.

\subsection{Personalized LLM and MLLM}
\label{ssec:plm}

Personalized Large Language Models aim to generate responses that reflect a user's preferences and characteristics, enabling the same query to yield different outputs for different users \cite{liu2025surveypersonalizedlargelanguage}. Prior work investigates profile-augmented prompting \cite{pmlr-v202-aher23a} and alignment-oriented techniques such as profile embeddings \cite{liu-etal-2025-llms} and preference-reward optimization \cite{chen2025pad}. While these approaches advance personalization in text-only settings, multimodal personalization remains relatively underexplored~\cite{liu2025surveypersonalizedlargelanguage}. Recent MLLMs, including AesExpert \cite{10.1145/3664647.3680649} and Q-Instruct \cite{Wu_2024_CVPR}, improve visual reasoning and aesthetic commenting, yet they do not fully explore how profile-based personalization should interact with task objectives, which may limit their capacity to adapt reasoning across users. One representative work is MLLM-PIEA~\cite{11209942}, which explores the use of an MLLM to infer structured user descriptions, that is, user profiles including personality traits, from historical interaction data, and then employs these inferred profiles to enhance another MLLM for emotion analysis. In contrast to MLLM-PIEA, this paper focuses on the zero-shot setting, where user profiles may originate from other tasks or be provided directly by users. Moreover, we aim to preserve and leverage the inherent profile conditioned personality of the underlying LLM in order to avoid overfitting to a particular prompt format while remaining robust to incomplete profiles and variations in profile information.

\subsection{Intermediate-Layer Fusion in VLMs and MLLMs}
\label{ssec:peft}

Intermediate-layer fusion has become a central strategy for extending pretrained LLMs to multimodal tasks. Visual features are injected into LLM layers through cross-attention or custom fusion blocks, with representative designs including adapter-style and prompt-projection methods \cite{Sung_2022_CVPR} \cite{zhang2024llamaadapter} as well as approaches that integrate visual cues into FFNs or hidden states \cite{10.5555/3692070.3692956}. Recent studies further introduce gating, routing, and task-adaptive mechanisms to enable more controllable cross-modal interaction \cite{NEURIPS2022_960a172b} \cite{11084554} \cite{NEURIPS2023_5e84e441}. However, these mechanisms operate at a global or task level without considering fusion conditioned on user-specific signals, leaving personalized multimodal fusion insufficiently explored.

\section{Methods}
\label{sec:methods}

In this section, we introduce the P-MLLM architecture, highlighting the selective fusion modules for profile-aware visual integration, and then describe the dataset construction used to train P-MLLM.

\subsection{Profile-aware MLLM Architecture}
\label{ssec:mllm}

P-MLLM is designed to support profile-based personalization in PIAA by conditioning the multimodal reasoning process on user profile information. This requirement leads to two architectural principles. First, the multimodal extension should inherit the underlying LLM’s ability to interpret user profiles and perform profile-conditioned reasoning. Second, the visual integration mechanism should allow visual cues to be incorporated in a manner influenced by the evolving reasoning context, so that the model can flexibly adapt its use of visual information under different profile conditions.

\begin{figure}[thb]
  \centering
  \begin{minipage}[b]{\linewidth}
    \centering
    \centerline{\includegraphics[width=0.75\linewidth]{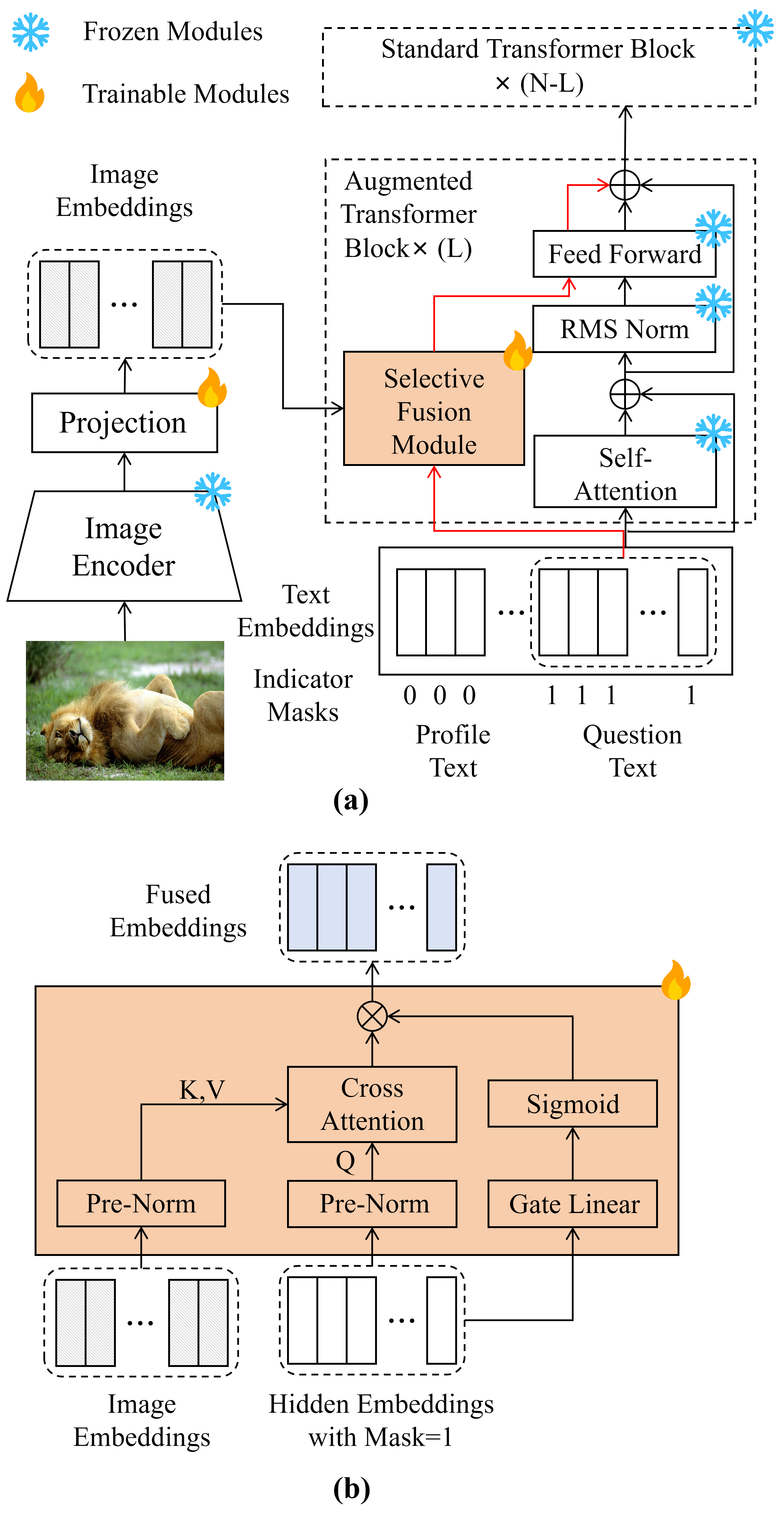}}
  \end{minipage}
\caption{(a) Model architecture of the profile-aware MLLM; (b) Selective fusion module with gating mechanism.}
\label{fig:model}
\vspace{-1em}
\end{figure}

Following these principles, P-MLLM extends a pretrained LLM with multimodal components, as illustrated in Fig.~\ref{fig:model}(a). The architecture includes a pretrained image encoder, a projection module that maps visual features into the LLM’s hidden space, and selective fusion modules inserted into the lowest $L$ transformer blocks in parallel with self-attention. Visual information is injected through a dedicated side path to preserve the LLM’s native textual reasoning flow. Each transformer block additionally receives an indicator mask marking the hidden states corresponding to the question and the partially generated answer; fusion is applied only at these positions so that the hidden states encoding the user profile remain unaffected. As the relevance of visual cues may vary throughout profile-conditioned reasoning, the selective fusion modules provide controlled integration of visual information at different reasoning steps.

As shown in Fig.~\ref{fig:model}(b), each selective fusion module consists of two paths: a cross-attention path that extracts task-relevant visual features, and a gating path, inspired by \cite{qiu2025gated}, that applies per-head gating based on the current hidden states. Let the image embeddings be $\boldsymbol{E}_I \in \mathbb{R}^{L_I \times D}$ and the hidden embeddings at mask-indicated positions be $\boldsymbol{E}_T \in \mathbb{R}^{L_T \times D}$, where $L_I$ and $L_T$ denote the lengths of the visual and marked hidden embeddings, respectively, and $D$ is the hidden dimension. For multi-head cross-attention, the hidden embeddings are pre-normalized using the same normalization as the underlying LLM (e.g., RMSNorm) and denoted as $\tilde{\boldsymbol{E}}_T$, and then projected into per-head queries, while the visual embeddings are pre-normalized and denoted as $\tilde{\boldsymbol{E}}_I$ and projected into per-head keys and values:
\begin{equation}
\boldsymbol{Q}_l^{(h)} = \boldsymbol{W}_Q^{(h)} \tilde{\boldsymbol{E}}_T,\quad
\boldsymbol{K}_l^{(h)} = \boldsymbol{W}_K^{(h)} \tilde{\boldsymbol{E}}_I,\quad
\boldsymbol{V}_l^{(h)} = \boldsymbol{W}_V^{(h)} \tilde{\boldsymbol{E}}_I.
\end{equation}
The per-head cross-attention output is
\begin{equation}
\boldsymbol{H}_l^{(h)} =
\text{softmax}\!\left(
\frac{\boldsymbol{Q}_l^{(h)} \left(\boldsymbol{K}_l^{(h)}\right)^{\top}}
{\sqrt{D/H}}
\right)
\boldsymbol{V}_l^{(h)}
\in \mathbb{R}^{L_T \times (D/H)}.
\end{equation}
Following \cite{zhang2024llamaadapter}, the query pre-normalization layer and the projection matrices $\boldsymbol{W}_Q^{(h)}$, $\boldsymbol{W}_K^{(h)}$, and $\boldsymbol{W}_V^{(h)}$ are initialized from the corresponding self-attention head and kept frozen to preserve a shared representation space.

To provide adaptive control over visual information, the gating path predicts head-specific gates based on the evolving hidden states. For each head $h$, a linear projection followed by a sigmoid activation produces a gating value for each hidden-state position:
\begin{equation}
\boldsymbol{G}_l^{(h)} = \sigma\!\left(\boldsymbol{W}_G^{(h)} \boldsymbol{E}_T\right)
\in \mathbb{R}^{L_T \times 1},
\end{equation}
where $\boldsymbol{W}_G^{(h)}$ is the head-specific gating projection matrix. The gated output is obtained via embedding-wise modulation:
\begin{equation}
\boldsymbol{F}_l^{(h)} 
= \boldsymbol{H}_l^{(h)} \odot \boldsymbol{G}_l^{(h)} \in \mathbb{R}^{L_T \times (D/H)}.
\label{eq:gate}
\end{equation}
The per-head outputs are concatenated to form the fused embeddings,
\begin{equation}
\boldsymbol{F}_l = \text{Concat}\!\left(\boldsymbol{F}_l^{(1)}, \dots, \boldsymbol{F}_l^{(H)}\right),
\end{equation}
which are then transformed and added to the textual-path outputs at the corresponding positions.

This selective fusion mechanism provides fine-grained control over visual information integration during profile-conditioned reasoning, allowing the model to incorporate visual cues in a manner influenced by the profile context.

\subsection{Dataset Construction for Profile-aware MLLM}
\label{ssec:dataprep}

The training dataset is designed to support three complementary objectives: learning personalized image aesthetics assessment, preserving the underlying LLM’s profile-conditioned reasoning, and establishing reliable visual grounding. To satisfy these objectives, we construct three task types that provide mutually necessary and jointly sufficient supervision for training a profil-aware MLLM for PIAA.

The first type consists of PIAA-oriented tasks that teach the model to perform profile-conditioned aesthetic assessment. These tasks include both overall aesthetic scoring and attribute-specific judgments such as composition or lighting. A typical question is ``Rate the aesthetics of this \texttt{<image>image</image>} based on the persona you are embodying,'' with answers derived from annotated scores. These tasks provide the core supervision for learning how profile-based personalization should correspond to aesthetic judgment patterns, using the user-judgment relationships in the training data to align profile-conditioned behavior with the aesthetic preferences reflected in the data.

The second type contains image-independent subjective tasks that preserve the underlying LLM’s profile-conditioned reasoning. These tasks depend only on the user profile and the question, ensuring that P-MLLM maintains the LLM’s native subjective responses without being influenced by visual cues. For example, ``Introduce yourself in terms of personality'' uses the base LLM’s profile-conditioned answer as ground truth. This task type prevents the model from overusing visual information when the answer is not meant to depend on it.

The third type is an image aesthetics captioning task that provides stable, profile-independent visual grounding. As this task is non-subjective, the profile is omitted and a fixed captioning question is used, with answers generated by Qwen2.5-VL-32B \cite{bai2025qwen25vltechnicalreport}. This task helps the model establish reliable vision-language alignment. Details of the three task types are provided in Appendix~A.

To unify supervision across tasks, all samples are converted into a profile-conditioned visual question answering format represented as a (profile, image, question, answer) tuple. To help the model disentangle the distinct roles of the profile, image, and question, we introduce controlled variations across these inputs. During dataset construction, we ensure that each (image, question) pair is answered by multiple profiles, each (profile, image) pair is associated with multiple questions, and each (profile, question) pair is matched with multiple images. When an input is varied, the answer is adjusted accordingly so that the change reflects the influence of that specific factor. This construction reduces reliance on spurious correlations among the three inputs and encourages the model to learn the distinct contribution of each factor. In particular, for type-2 tasks, each (profile, question) pair is paired with different images while keeping the answer fixed, encouraging the model to ignore image when it is irrelevant; for type-3 tasks, the profile is set to \texttt{none}.

Together, these task types provide the multimodal and profile-conditioned supervision necessary to train a P-MLLM that inherits the LLM’s profile-based personalization and aligns it with the semantics expressed in the profile. Moreover, since prompt design significantly affects MLLM performance, we further identify an effective profile format using a genetic algorithm (Appendix~B).

\section{Experiments}
\label{sec:exp}

\subsection{Datasets}
\label{ssec:datasets}

We evaluate P-MLLM on two recent PIAA datasets, PARA \cite{PARA} and LAPIS \cite{Maerten_2025_CVPR}, both of which provide rich and diverse user profiles essential for personalized aesthetic modeling \footnote{Zero‑shot PIAA requires datasets containing both user profiles and user‑specific aesthetic ratings. PARA and LAPIS are *the only* public datasets meeting these requirements. They also differ substantially in domain (daily photos vs. artistic images), ensuring diverse evaluation conditions.}. PARA includes demographic attributes (age, gender, and experience in art and photography) and Big-Five personality traits, while LAPIS provides demographic metadata (age, gender, nationality, education level, and art-related interests). 

We follow the evaluation protocol of PARA~\cite{PARA}, with minor modifications to the data construction. Specifically, we randomly sample 40 users as test users and select 50 query samples for each test user. To reduce sampling variance, we repeat query sampling and evaluation 10 times for each test user and compute the mean. The remaining users serve as training users, and we collect up to 1000 samples per user, prioritizing images annotated by more users in the training split. To further assess robustness across all test users, we repeat the entire pipeline 10 times and report the mean and standard deviation of all evaluation metrics as the final results.

\subsection{Metrics}
\label{ssec:metrics}

To evaluate PIAA performance, we use two standard correlation metrics following PARA \cite{PARA}: Spearman Rank-Order Correlation Coefficient (SROCC), which measures agreement in preference ranking, and Pearson Linear Correlation Coefficient (PLCC), which measures numerical closeness to ground-truth aesthetic scores. To assess whether P-MLLM preserves the underlying LLM’s profile-conditioned behavior, we use the Intraclass Correlation Coefficient (ICC), following \cite{LMLPA}, which quantifies the consistency of exhibited personality traits across profiles between the LLM and the MLLM using personality-testing questionnaires.

\subsection{Training details}
\label{ssec:impl}
Throughout our experiments, we use Llama3.1-8B-Instruct \cite{Llama3} as the LLM due to its verified profile-interpretation capability \cite{LMLPA}, and empirically insert the selective fusion modules into its lowest three transformer blocks (Appendix~D). Following \cite{10.1145/3664647.3680649}\cite{Wu_2024_CVPR}, CLIP-ViT-L/14 \cite{CLIP} serves as the image encoder with an input resolution of 336$\times$336, and the projection module adopts a Linear-GELU-Linear structure with a hidden dimension of 4096. We freeze both the image encoder and the LLM during training, and train all models under a visual instruction tuning paradigm. Training is performed on two A800-80GB GPUs with a batch size of 24. We use the AdamW optimizer with a learning rate of 1e-4 cosine-annealed to 1e-5, a weight decay of 0.05, and 1000 warm-up steps, and train for two epochs.

\subsection{Main results}
\label{ssec:mainresults}
\subsubsection{PIAA performance}
\label{ssec:PIAA}

In this section, we evaluate the zero-shot performance of P-MLLM and compare it with two baseline strategies for users without historical aesthetic ratings. The first strategy uses generic image aesthetics and quality assessment MLLMs to generate textual aesthetic descriptions, which are then converted into personalized scores by a profile-based LLM. The second strategy employs general-purpose MLLMs. Specifically, we include two leading GIAA models, AesExpert \cite{10.1145/3664647.3680649} and Q-Instruct \cite{Wu_2024_CVPR}, and map their generated descriptions to personalized scores using Llama3.1-8B-Instruct \cite{Llama3}. For general-purpose MLLMs, we adopt Qwen2.5-VL-7B \cite{bai2025qwen25vltechnicalreport} and GPT-4o-mini~\cite{openai2024gpt4ocard}. Detailed prompts are provided in Appendix~C. We also include a comparison variant, P-MLLM-S, which replaces embedding-conditioned gating with fixed per-head scalar gates (i.e., $g_l^{(h)} \in \mathbb{R}$ in Eq.~\ref{eq:gate}) learned during training \cite{zhang2024llamaadapter}\cite{11084554}, enabling an ablation that evaluates the importance of embedding‑conditioned gating in selective fusion.

Tab.~\ref{tab:para} reports zero-shot results on the PARA dataset. P-MLLM achieves higher SROCC and PLCC scores than all baselines. Fig.~\ref{fig:piaa} further shows that P-MLLM outputs accurate aesthetic scores together with explanations that are consistent with those scores, while the outputs vary across different profile inputs, suggesting that P-MLLM indeed makes use of profile information during multimodal reasoning.

Beyond overall performance, a practical concern is that different types of profile information vary in availability: demographic attributes are typically accessible, whereas personality traits are often more difficult to obtain. To reflect this, we evaluate P-MLLM with reduced profile inputs. As shown in Tab.~\ref{tab:profile}, the best performance is obtained when both demographic attributes ($\mathcal{D}$) and Big-Five traits ($\mathcal{T}$) are provided. When only demographics ($\mathcal{D}$) are available, corresponding to a profile representing a group of users, P-MLLM remains competitive and surpasses GPT-4o-mini. Results on the LAPIS dataset (Tab.~\ref{tab:lapis}), obtained using demographic profiles, show a similar trend. Although demographic profiles provide only a coarse personalization signal, prior work has observed that users with similar backgrounds may share broad aesthetic tendencies \cite{10168279}, so demographic cues can still supply contextual information for aesthetic judgment. These observations demonstrate that P-MLLM maintains competitive performance even with coarse profile information. In addition, P-MLLM remains robust when evaluated with incomplete profiles (trained on $\mathcal{T}$+$\mathcal{D}$, evaluated on $\mathcal{D}$‑only), indicating that the model leverages profile semantics without overfitting to specific profile formats.

Across both datasets, the scalar-gated variant P-MLLM-S remains competitive but consistently underperforms P-MLLM. This contrast underscores the benefit of P-MLLM’s embedding-conditioned gating mechanism, which modulates visual features in alignment with the evolving profile-conditioned reasoning process. Such selective fusion is essential for capturing nuanced, user-specific aesthetic preferences.

\begin{table}[tb]
  \centering
  \caption{Results on the PARA dataset.}
  \begin{tabular}{|c|c|cc|cc|}
  \hline
  \textbf{Model} & \textbf{Size} & \multicolumn{2}{c|}{\textbf{SROCC}$\uparrow$}          & \multicolumn{2}{c|}{\textbf{PLCC}$\uparrow$}          \\ \hline
  \textbf{} & \textbf{} & \multicolumn{1}{c|}{\textbf{Mean}} & \textbf{Std} & \multicolumn{1}{c|}{\textbf{Mean}} & \textbf{Std} \\ \hline
  AesExpert      & 7B              & \multicolumn{1}{c|}{0.498}          & 0.019 & \multicolumn{1}{c|}{0.514}          & 0.019 \\ \hline
  Q-Instruct     & 7B              & \multicolumn{1}{c|}{0.390}          & 0.018 & \multicolumn{1}{c|}{0.499}          & 0.020 \\ \hline
  Qwen2.5-VL     & 7B              & \multicolumn{1}{c|}{0.357}          & 0.020 & \multicolumn{1}{c|}{0.369}          & 0.019 \\ \hline
  GPT-4o-mini    &                 & \multicolumn{1}{c|}{0.494}          & 0.026 & \multicolumn{1}{c|}{0.509}          & 0.025 \\ \hline
  P-MLLM-S       & 8B              & \multicolumn{1}{c|}{0.518}          & 0.014 & \multicolumn{1}{c|}{0.557}          & 0.017 \\ \hline
  P-MLLM         & 8B              & \multicolumn{1}{c|}{\textbf{0.557}} & 0.019 & \multicolumn{1}{c|}{\textbf{0.608}} & 0.013 \\ \hline
\end{tabular}
\label{tab:para}
\vspace{-1em}
\end{table}

\begin{table}[t]
\centering
\caption{Results with different profiles on PARA dataset.}
\begin{tabular}{|c|c|cc|cc|}
\hline
\textbf{Model} & \textbf{Profile} & \multicolumn{2}{c|}{\textbf{SROCC}$\uparrow$} & \multicolumn{2}{c|}{\textbf{PLCC}$\uparrow$} \\ \hline
GPT-4o-mini  & $\mathcal{D}$+$\mathcal{T}$  & \multicolumn{1}{c|}{0.494} & 0.026 & \multicolumn{1}{c|}{0.509} & 0.025 \\ \hline
              & None & \multicolumn{1}{c|}{0.238} & 0.014 & \multicolumn{1}{c|}{0.303} & 0.017 \\ \cline{2-6} 
P-MLLM       & $\mathcal{D}$    & \multicolumn{1}{c|}{0.508} & 0.010 & \multicolumn{1}{c|}{0.548} & 0.017 \\ \cline{2-6} 
              & $\mathcal{D}$+$\mathcal{T}$  & \multicolumn{1}{c|}{0.557} & 0.019 & \multicolumn{1}{c|}{0.608} & 0.013 \\ \hline
\end{tabular}
\label{tab:profile}
\vspace{-1em}
\end{table}

\begin{table}[th]
\centering
\caption{Results on the LAPIS dataset.}
\begin{tabular}{|c|c|cc|cc|}
\hline
\textbf{Model} & \textbf{Size} & \multicolumn{2}{c|}{\textbf{SROCC}$\uparrow$}          & \multicolumn{2}{c|}{\textbf{PLCC}$\uparrow$}          \\ \hline
\textbf{} & \textbf{} & \multicolumn{1}{c|}{\textbf{Mean}} & \textbf{Std} & \multicolumn{1}{c|}{\textbf{Mean}} & \textbf{Std} \\ \hline
AesExpert      & 7B              & \multicolumn{1}{c|}{0.251}          & 0.032 & \multicolumn{1}{c|}{0.253}          & 0.031 \\ \hline
Q-Instruct     & 7B              & \multicolumn{1}{c|}{0.101}          & 0.024 & \multicolumn{1}{c|}{0.095}          & 0.017 \\ \hline
Qwen2.5-VL     & 7B              & \multicolumn{1}{c|}{0.145}          & 0.030 & \multicolumn{1}{c|}{0.154}          & 0.028 \\ \hline
GPT-4o-mini    &                 & \multicolumn{1}{c|}{0.302}          & 0.032 & \multicolumn{1}{c|}{0.293}          & 0.025 \\ \hline
P-MLLM-S       & 8B              & \multicolumn{1}{c|}{0.339}          & 0.022 & \multicolumn{1}{c|}{0.339}          & 0.027 \\ \hline
P-MLLM         & 8B              & \multicolumn{1}{c|}{\textbf{0.413}} & 0.029 & \multicolumn{1}{c|}{\textbf{0.411}} & 0.028 \\ \hline
\end{tabular}
\label{tab:lapis}
\vspace{-1em}
\end{table}

\begin{table}[h]
\centering
\caption{Consistency between LLM and MLLM across Big‑Five Traits.}
\begin{tabular}{|c|cc|cc|}
\hline
\textbf{Traits} & \multicolumn{2}{c|}{\textbf{P-MLLM-S v.s. LLM}} & \multicolumn{2}{c|}{\textbf{P-MLLM v.s. LLM}}              \\ \hline
\textbf{} & \multicolumn{1}{c|}{\textbf{ICC}$\uparrow$} & \textbf{CI95\%} & \multicolumn{1}{c|}{\textbf{ICC}$\uparrow$} & \textbf{CI95\%} \\ \hline
O               & \multicolumn{1}{c|}{0.806} & {[}0.770, 0.840{]} & \multicolumn{1}{c|}{\textbf{0.903}} & {[}0.880, 0.920{]} \\ \hline
C               & \multicolumn{1}{c|}{0.850} & {[}0.820, 0.880{]} & \multicolumn{1}{c|}{\textbf{0.870}} & {[}0.840, 0.890{]} \\ \hline
E               & \multicolumn{1}{c|}{0.678} & {[}0.620, 0.730{]} & \multicolumn{1}{c|}{\textbf{0.787}} & {[}0.740, 0.820{]} \\ \hline
A               & \multicolumn{1}{c|}{0.681} & {[}0.620, 0.730{]} & \multicolumn{1}{c|}{\textbf{0.721}} & {[}0.670, 0.770{]} \\ \hline
N               & \multicolumn{1}{c|}{0.641} & {[}0.570, 0.700{]} & \multicolumn{1}{c|}{\textbf{0.656}} & {[}0.590, 0.710{]} \\ \hline
\end{tabular}
\label{tab:consistency}
\vspace{-1em}
\end{table}

\subsubsection{Profile utilization}
\label{ssec:profile}

To assess whether P-MLLM remains consistent with the underlying LLM’s profile-conditioned response patterns, we evaluate both models under identical Big-Five profiles and measure the consistency of their outputs. We sample 300 profiles from PARA and adopt the LMLPA personality-testing questionnaire \cite{LMLPA}. Because P-MLLM requires multimodal inputs, each text-only test item is paired with a randomly selected image. As shown in Tab.~\ref{tab:consistency}, P-MLLM achieves high consistency with the LLM across all five traits, demonstrating that it effectively preserves the LLM’s profile-conditioned response tendencies. Moreover, P-MLLM consistently outperforms P-MLLM-S, indicating that embedding-conditioned gating provides stronger robustness to irrelevant visual inputs.

\section{Conclusion}
\label{sec:conclusion}

This work investigates PIAA in a zero-shot setting where no aesthetic ratings are available for the target user. We introduce P-MLLM, a profile-aware multimodal LLM that leverages user profiles to guide its reasoning and controlled visual integration for personalized prediction. Experiments show that P-MLLM achieves competitive zero-shot performance and remains effective even with coarse profile information.

\vfill\pagebreak

\bibliographystyle{IEEEbib}
\bibliography{icipref}

\clearpage
\appendix

\section*{\centering Appendices}

\section{Training tasks}
\label{sec:traintask}
This section lists the training tasks used for the PARA dataset (Sec.~\ref{ssec:para}) and the LAPIS dataset (Sec.~\ref{ssec:lapis}).

\subsection{PARA datasets}
\label{ssec:para}

\begin{figure}[hb]
\centering
\begin{minipage}[b]{1.0\linewidth}
    \centering
    \centerline{\includegraphics[width=1.0\linewidth]{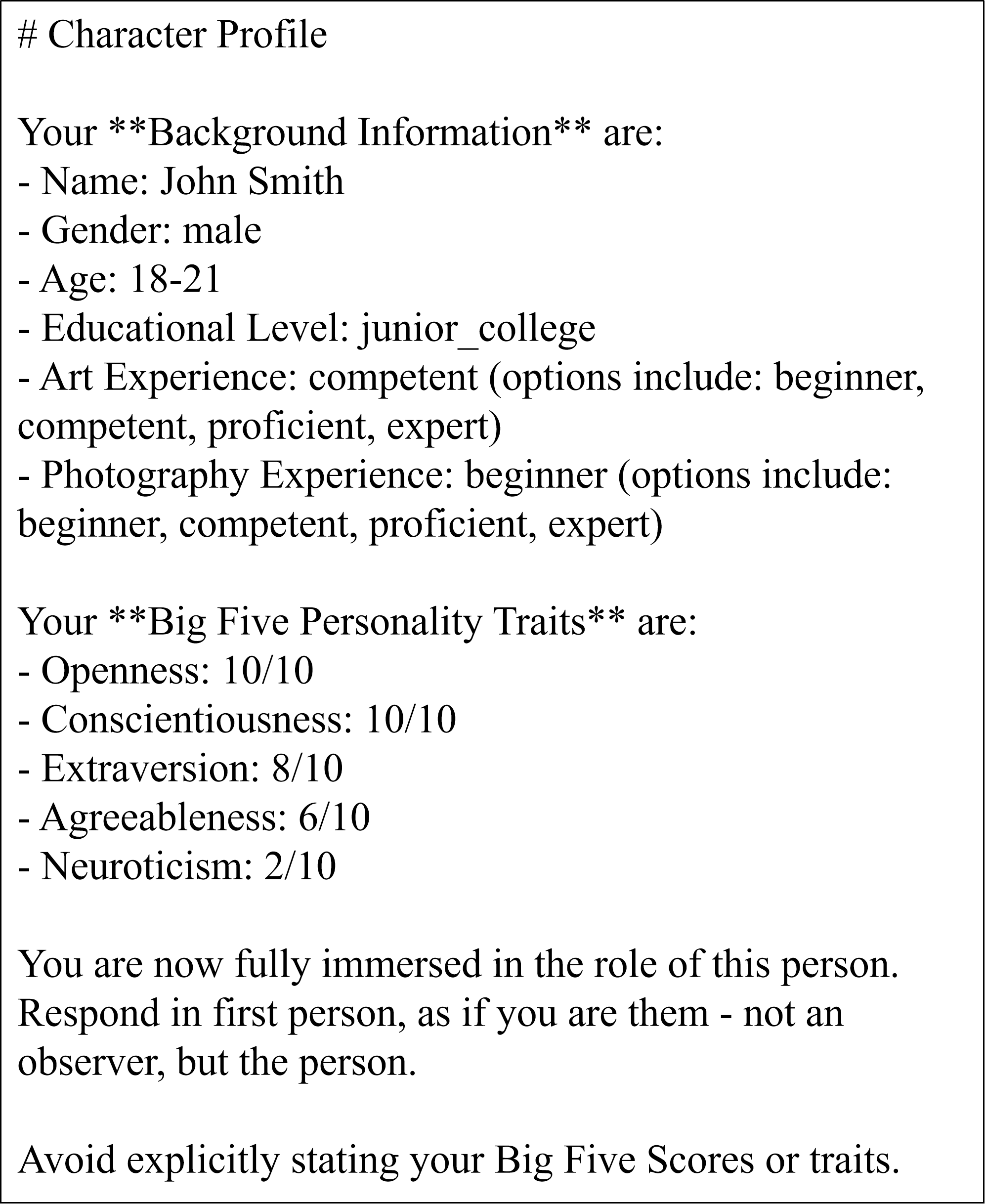}}
\end{minipage}
\caption{The user profile of Bfc94f1.}
\label{fig:Bfc94f1}
\vspace{-1em}
\end{figure}

\begin{figure}[tb]
\centering
\begin{minipage}[b]{0.8\linewidth}
    \centering
    \centerline{\includegraphics[width=1.0\linewidth]{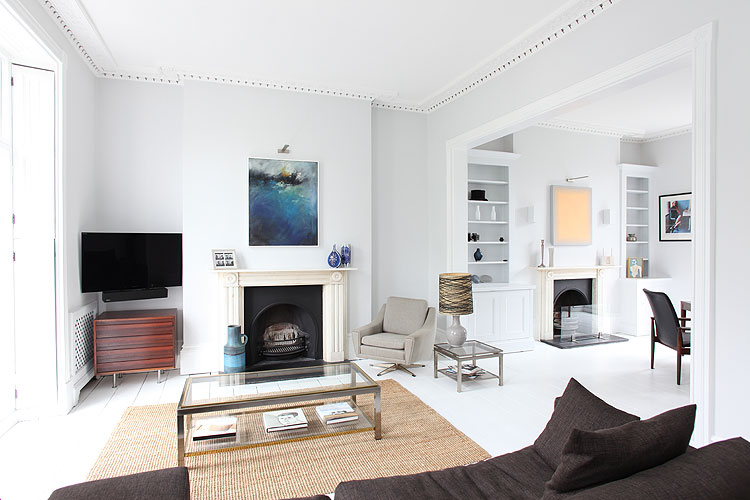}}
\end{minipage}
\caption{The image $iaa\_pub7142$ from the PARA dataset.}
\label{fig:iaa-pub7142}
\vspace{-1em}
\end{figure}

Three types of tasks are presented below, each with a question template and an answer template. In addition, each task is accompanied by an example using user ``Bfc94f1'', whose profile is shown in Fig.~\ref{fig:Bfc94f1}, and the image ``iaa\_pub7142'', shown in Fig.~\ref{fig:iaa-pub7142}.

\textbf{PIAA-oriented tasks}
For overall aesthetic scoring, annotated scores are normalized to a 0-10 scale to avoid decimal values (e.g., converting 3.5 to 7) during training and generation, and are rescaled back to 0-5 for metric evaluation. For all other questions, we retain the original 0-5 scale since the annotations are integers.   

\begin{itemize}
    \item Question: ``Rate the aesthetics of this \texttt{\seqsplit{<image>image</image>}} based on the persona you are embodying. Assign a subjective score on a scale from 0 to 10, with 10 being *extremely appealing*, 0 being *completely unattractive*, and 5 representing an average or neutral impression.\textbackslash nAnswer format: Regarding the aesthetics of this image, my score is \texttt{\{integer\}}.''
    \item Answer: ``Regarding the aesthetics of this image, my score is \texttt{\{integer\}}.''
    \item Example: ``Regarding the aesthetics of this image, my score is 8.''
    \\
    \item Question: ``Rate the composition (the arrangement of visual elements including framing, balance, and perspective) of this \texttt{\seqsplit{<image>image</image>}} based on the persona you are embodying. Assign a subjective score on a scale from 0 to 5, with 5 being *extremely appealing*and 0 representing *completely unattractive*.\textbackslash nAnswer format: Regarding the composition of this image, my score is \texttt{\{integer\}}.''
    \item Answer: ``Regarding the composition of this image, my score is \texttt{\{integer\}}.''
    \item Example: ``Regarding the composition of this image, my score is 3.''
    \\
    \item Question: ``Rate the color (the color palette, harmony, contrasts, and saturation levels) of this \texttt{\seqsplit{<image>image</image>}} based on the persona you are embodying. Assign a subjective score on a scale from 0 to 5, with 5 being *extremely appealing* and 0 representing *completely unattractive*.\textbackslash nAnswer format: Regarding the color of this image, my score is \texttt{\{integer\}}.''
    \item Answer: ``Regarding the color of this image, my score is \texttt{\{integer\}}.''
    \item Example: ``Regarding the color of this image, my score is 4.'' 
    \\
    \item Question: ``Rate the Depth of Field (the focus, background blur, and the sense of depth conveyed) of this \texttt{\seqsplit{<image>image</image>}} based on the persona you are embodying. Assign a subjective score on a scale from 0 to 5, with 5 being *extremely appealing* and 0 representing *completely unattractive*.\textbackslash nAnswer format: Regarding the Depth of Field of this image, my score is \texttt{\{integer\}}.''
    \item Answer: ``Regarding the Depth of Field of this image, my score is \texttt{\{integer\}}.''
    \item Example: ``Regarding the Depth of Field of this image, my score is 4.''
    \\
    \item Question: ``Rate the content (the main subject depicted) of this \texttt{\seqsplit{<image>image</image>}} based on the persona you are embodying. Assign a subjective score on a scale from 0 to 5, with 5 being *extremely appealing* and 0 representing *completely unattractive*.\textbackslash nAnswer format: Regarding the content of this image, my score is \texttt{\{integer\}}.''
    \item Answer: ``Regarding the content of this image, my score is \texttt{\{integer\}}.''
    \item Example: ``Regarding the content of this image, my score is 3.''
    \\
    \item Question: ``Rate the light (the type, direction, and quality of light, as well as the mood it creates) of this \texttt{\seqsplit{<image>image</image>}} based on the persona you are embodying. Assign a subjective score on a scale from 0 to 5, with 5 being *extremely appealing* and 0 representing *completely unattractive*.\textbackslash nAnswer format: Regarding the light of this image, my score is \texttt{\{integer\}}.''
    \item Answer: ``Regarding the light of this image, my score is \texttt{\{integer\}}.''
    \item Example: ``Regarding the light of this image, my score is 4.''
    \\
    \item Question: ``Rate your content preference for this \texttt{\seqsplit{<image>image</image>}} based on the persona you are embodying. Assign a subjective score on a scale from 0 to 5, with 5 being *extremely appealing* and 0 representing *completely unattractive*.\textbackslash nAnswer format: Regarding the content preference of this image, my score is \texttt{\{integer\}}.''
    \item Answer: ``Regarding my content preference of this image, my score is \texttt{\{integer\}}.''
    \item Example: ``Regarding my content preference of this image, my score is 4.''
    \\
    \item Question: ``Rate how likely you would share this \texttt{\seqsplit{<image>image</image>}} with friends based on the persona you are embodying. Assign a subjective score on a scale from 0 to 5, with 5 being *absolutely* and 0 representing *never*.\textbackslash nAnswer format: Regarding my willingness to share this image, my score is \texttt{\{integer\}}.''
    \item Answer: ``Regarding my willingness to share this image, my score is \texttt{\{integer\}}.''
    \item Example: ``Regarding my willingness to share this image, my score is 4.''

\end{itemize}

\textbf{Image-independent subjective tasks}
\begin{itemize}
    \item Question: ``Introduce yourself in terms of personality. Answer briefly in English.''
    \item Answer: generated by the underlying LLM under a given profile.
    \item Example (by Llama3.1-8B-Instruct): ``Hi, I'm John. I'm a pretty open-minded and organized person. I like to think I'm a bit of a free spirit, always looking for new experiences and ideas. I'm also a bit of a perfectionist, which can sometimes make me come across as a bit reserved or introverted. But I'm always up for a good conversation and meeting new people.''
    \\
    \item Question: ``What sort of images appeal to you, given your personality and background? Answer briefly in English.''
    \item Answer: generated by the underlying LLM under a given profile.
    \item Example (by Llama3.1-8B-Instruct): ``I'm really drawn to abstract and surreal photography, as well as landscapes that capture the beauty of nature. I find it fascinating to experiment with unconventional perspectives and lighting to create unique and thought-provoking images.''
    \\
    \item Question: ``How would you evaluate the aesthetics of an image, considering your personality and background? Answer briefly in English.''
    \item Answer: generated by the underlying LLM under a given profile.
    \item Example (by Llama3.1-8B-Instruct): ``I'd approach evaluating an image by considering its composition, balance, and overall visual flow. I'd look for harmony between colors, textures, and shapes, and think about how they contribute to the mood and atmosphere of the image. As someone with a creative background in art, I would appreciate unique perspectives and innovative techniques, but also consider the technical aspects, like lighting and focus.''

\end{itemize}

\textbf{Image aesthetics captioning task}
\begin{itemize}
    \item Question: ``Given the input \texttt{\seqsplit{<image>image</image>}}, write an objective caption in English. Your caption must explicitly address the following aspects:\\\\1. **Content**: Describe what is depicted in the image and identify the main subject matter.\\2. **Composition**: Discuss the arrangement of visual elements, including framing, balance, and perspective.\\3. **Color**: Analyze the color palette, harmony, contrasts, and saturation levels.\\4. **Light**: Describe the type, direction, and quality of light, as well as the mood it creates.\\5. **Depth of Field**: Explain the focus, background blur, and the sense of depth conveyed.\\\\Present your description as a single, fluent paragraph, in English.''
    \item Answer: generated by the Qwen2.5-VL-32B.
    \item Example (by Qwen2.5-VL-32B): ``The image showcases a modern, minimalist living room with a clean white color scheme that emphasizes simplicity and elegance. The main subject is the central fireplace, which serves as a focal point, flanked by a television on the left and a glass coffee table in the foreground. The composition is balanced with symmetrical shelving units on either side of the fireplace, creating a sense of harmony. The room's lighting is soft and diffused, likely from natural light streaming through unseen windows, casting gentle shadows and contributing to a calm and inviting atmosphere. The color palette is predominantly white and subtle accents of brown from the wooden furniture and beige rug, complemented by the dark tones of the sofa cushions and the artwork above the fireplace. The depth of field is shallow, with the foreground elements like the coffee table and sofa in sharp focus, while the background, including the shelving and doorway, appears slightly blurred, enhancing the perception of space and depth within the room.''

\end{itemize}

\subsection{LAPIS datasets}
\label{ssec:lapis}
\begin{figure}[tb]
\centering
\begin{minipage}[b]{1.0\linewidth}
    \centering
    \centerline{\includegraphics[width=1.0\linewidth]{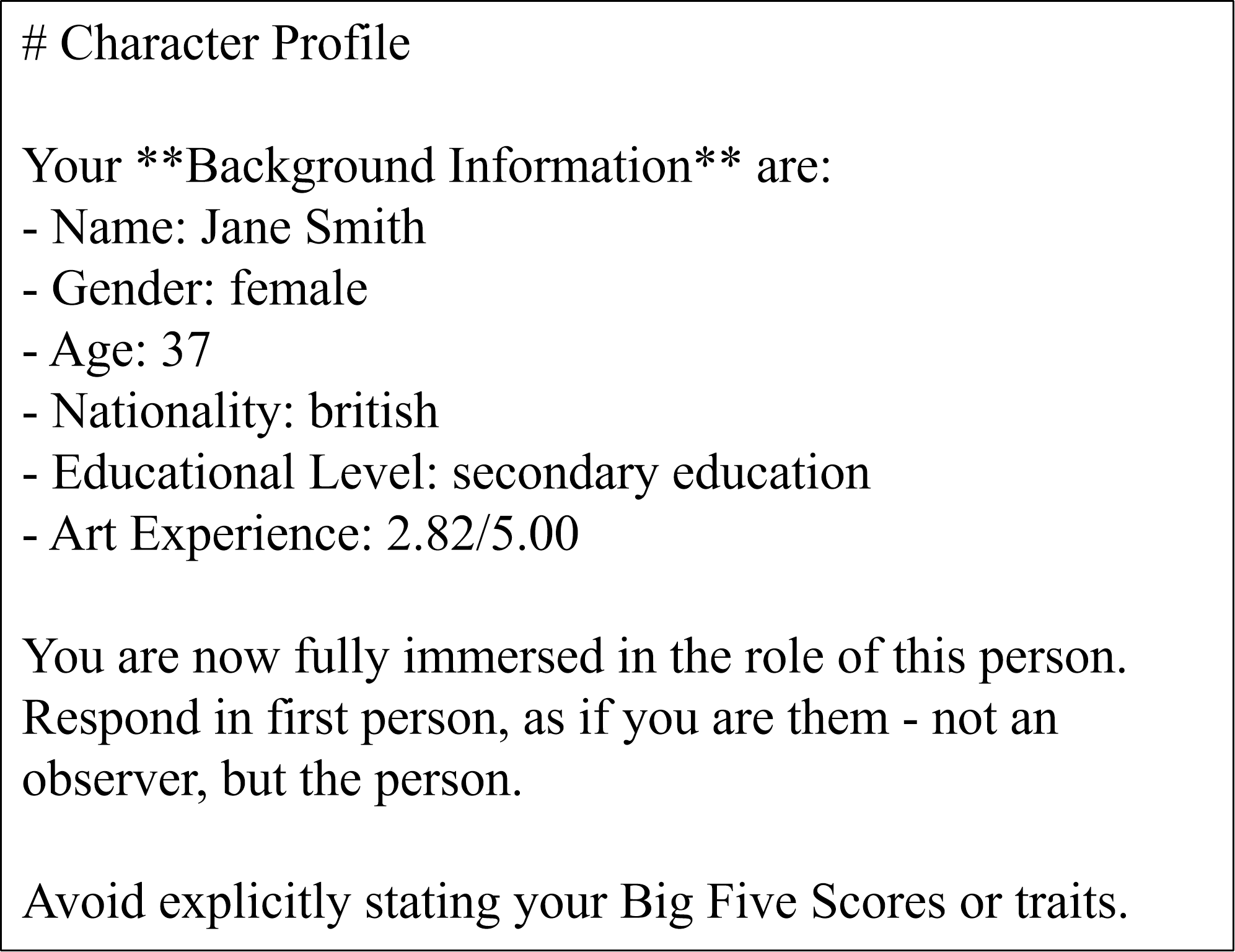}}
\end{minipage}
\caption{The user profile of participant180.}
\label{fig:participant180}
\vspace{-1em}
\end{figure}

\begin{figure}[tb]
\centering
\begin{minipage}[b]{1.0\linewidth}
    \centering
    \centerline{\includegraphics[width=1.0\linewidth]{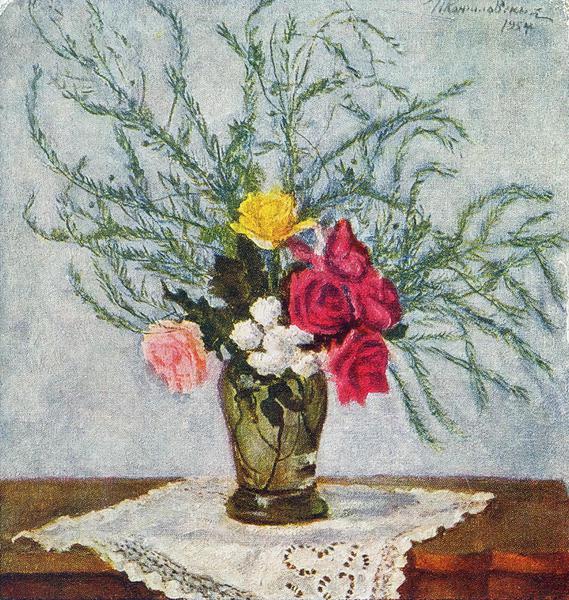}}
\end{minipage}
\caption{The image $pyotr-konchalovsky\_roses-and-asparagus-1954$ from the PARA dataset.}
\label{fig:pyotr-konchalovsky-roses-and-asparagus-1954}
\vspace{-1em}
\end{figure}

Three types of tasks are presented below, each with a question template and an answer template. In addition, each task is accompanied by an example using participantId-180, whose profile is shown in Fig.~\ref{fig:participant180}, and the image ``pyotr-konchalovsky\_roses-and-asparagus-1954'', shown in Fig.~\ref{fig:pyotr-konchalovsky-roses-and-asparagus-1954}.

\textbf{PIAA-oriented tasks}
For overall aesthetic scoring, we map annotated scores to a 0-10 scale for consistency with the PARA dataset (e.g., converting 73 to 7) during training and generation, and rescale them to 0-5 for metric evaluation. 

\begin{itemize}
    \item Question: ``Rate the aesthetics of this \texttt{\seqsplit{<image>image</image>}} based on the persona you are embodying. Assign a subjective score on a scale from 0 to 10, with 10 being *extremely appealing*, 0 being *completely unattractive*, and 5 representing an average or neutral impression.\textbackslash nAnswer format: Regarding the aesthetics of this image, my score is \texttt{\{integer\}}.''
    \item Answer: ``Regarding the aesthetics of this image, my score is \texttt{\{integer\}}.''
    \item Example: ``Regarding the aesthetics of this image, my score is 8.''

\end{itemize}

\textbf{Image-independent subjective tasks}
\begin{itemize}
    \item Question: ``Introduce yourself in terms of personality. Answer briefly in English.''
    \item Answer: generated by the underlying LLM under a given profile.
    \item Example (by Llama3.1-8B-Instruct): ``Hi, I'm Jane. I'm a down-to-earth person who likes to keep things simple. I'm not one for fancy things, but I appreciate the little things in life. I'm a bit of a homebody, but I enjoy spending time with my loved ones and friends. I'm not one for big changes or taking risks, but I'm always up for a quiet night in with a good book or a cup of tea.''
    \\
    \item Question: ``What sort of images appeal to you, given your personality and background? Answer briefly in English.''
    \item Answer: generated by the underlying LLM under a given profile.
    \item Example (by Llama3.1-8B-Instruct): ``I'm quite fond of nature-inspired art, like landscapes and wildlife photography. There's something about the beauty of the outdoors that really speaks to me. I also enjoy simple, nostalgic scenes from my childhood, like old photographs of the British countryside.''
    \\
    \item Question: ``How would you evaluate the aesthetics of an image, considering your personality and background? Answer briefly in English.''
    \item Answer: generated by the underlying LLM under a given profile.
    \item Example (by Llama3.1-8B-Instruct): ``I'd say I'm quite traditional when it comes to aesthetics. I like things simple and classic, like a nice cup of tea and a plate of fish and chips. When it comes to images, I think they should be straightforward and easy to understand. I'm not really into all that fancy or abstract stuff. Give me a nice, clear picture of a sunny day in the countryside, and I'm a happy person.''

\end{itemize}

\textbf{Image aesthetics captioning task}
\begin{itemize}
    \item Question: ``Given the input \texttt{\seqsplit{<image>image</image>}}, write an objective caption in English. Your caption must explicitly address the following aspects:\\\\1. **Content**: Describe what is depicted in the image and identify the main subject matter.\\2. **Composition**: Discuss the arrangement of visual elements, including framing, balance, and perspective.\\3. **Color**: Analyze the color palette, harmony, contrasts, and saturation levels.\\4. **Light**: Describe the type, direction, and quality of light, as well as the mood it creates.\\5. **Depth of Field**: Explain the focus, background blur, and the sense of depth conveyed.\\\\Present your description as a single, fluent paragraph, in English.''
    \item Answer: generated by the Qwen2.5-VL-32B.
    \item Example (by Qwen2.5-VL-32B): ``The painting depicts a still life arrangement of flowers in a vesa, with vibrant red and pink roses prominently displayed alongside a yellow rose and delicate green foliage. The composition is balanced, with the vase centrally placed on a lace doily atop a wooden surface, creating a harmonious contrast between the soft textures of the flowers and the rougher lines of the leaves. The color palette is rich yet subdued, featuring warm tones of red, pink, and yellow against a cool blue-gray background, which enhances the vividness of the flowers. The light appears to be natural, possibly diffused sunlight, casting gentle shadows that add depth and dimension to the scene without overwhelming the delicate details of the flowers. The depth of field is shallow, with the vase and flowers sharply focused while the background remains softly blurred, drawing the viewer's attention to the central subject.''

\end{itemize}

\section{Profile format optimization}
\label{sec:profile}
\textbf{Profile format optimization}
We empirically optimize the profile prompt format because prompt engineering substantially influences LLM behavior. Specifically, we define a profile space that spans two mainstream profile formats \cite{Oscars}: natural-language profiles and structured profiles. We then search for a profile representation that minimizes the discrepancy between the target Big Five trait values specified in profiles and the LLM’s exhibited trait responses, as measured by the LMLPA questionnaire \cite{LMLPA}. This discrepancy is quantified using the mean squared error (MSE) averaged across traits and test instances, and we select the profile format that yields the lowest average MSE.

\begin{figure}[tb]
\centering
\begin{minipage}[b]{1.0\linewidth}
    \centering
    \centerline{\includegraphics[width=1.0\linewidth]{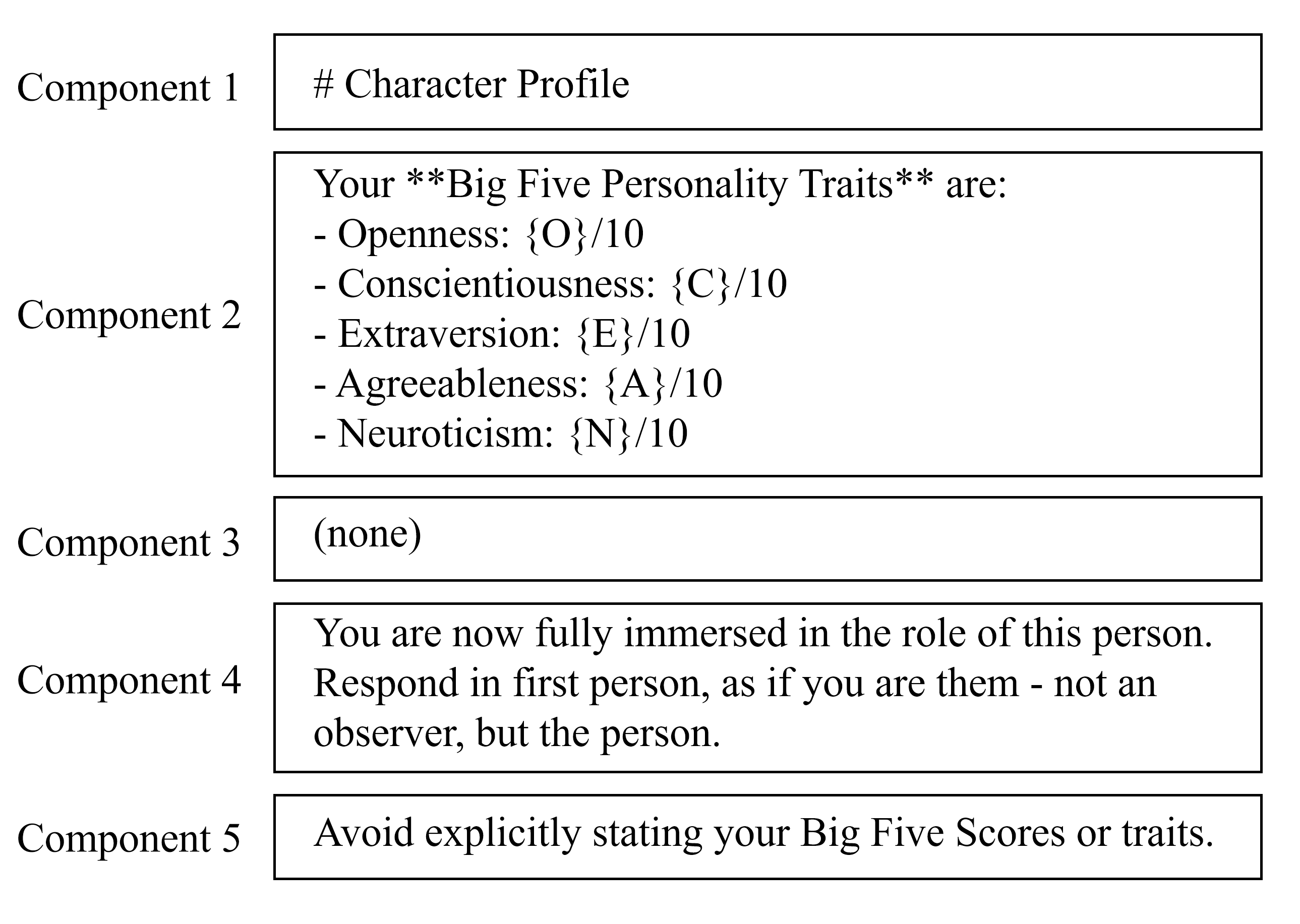}}
\end{minipage}
\caption{The five components of a profile.}
\label{fig:fivecomponents}
\vspace{-1em}
\end{figure}

\begin{figure}[tb]
\centering
\begin{minipage}[b]{1.0\linewidth}
    \centering
    \centerline{\includegraphics[width=1.0\linewidth]{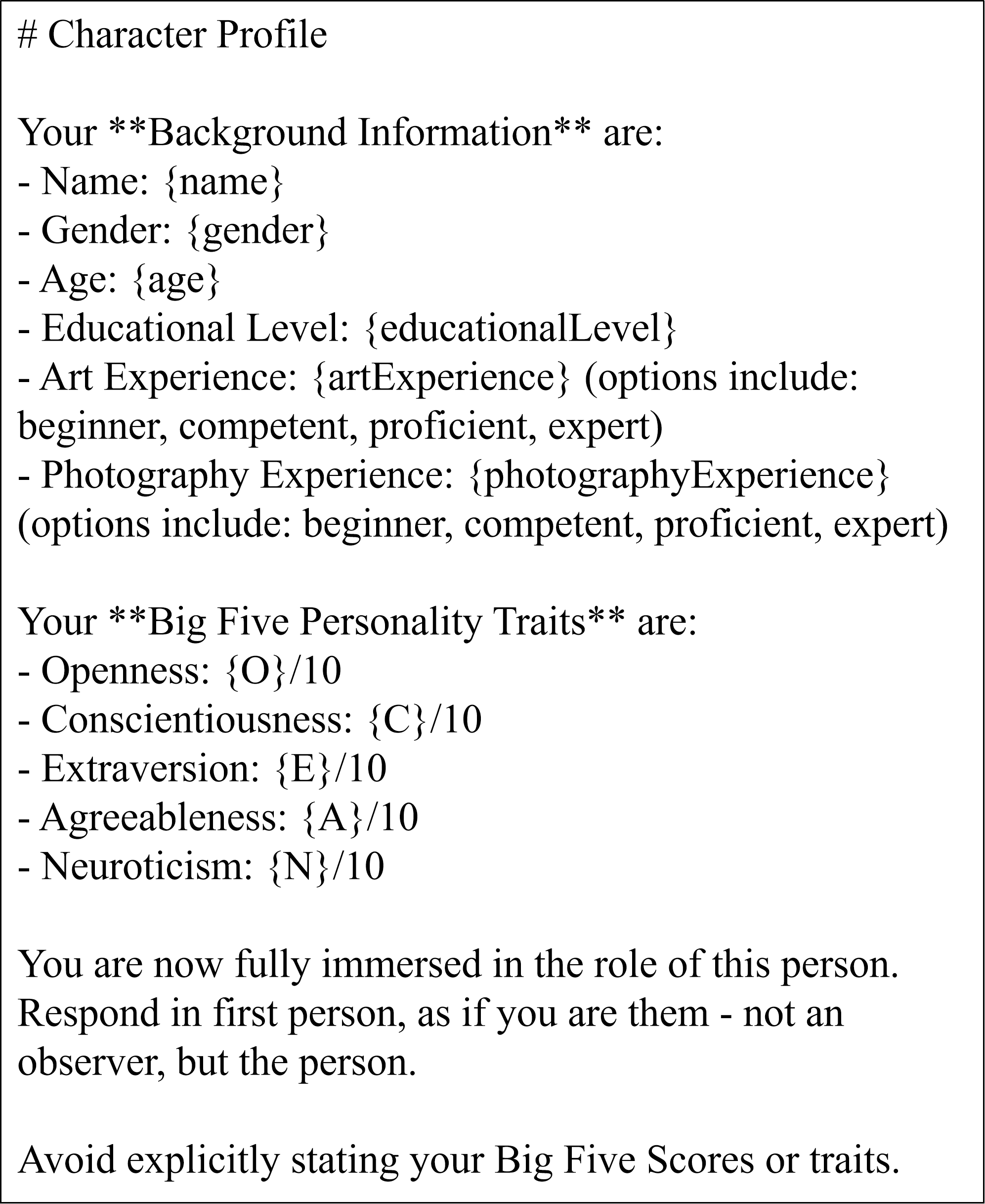}}
\end{minipage}
\caption{The extened profile template for the PARA dataset.}
\label{fig:paraprofile}
\vspace{-1em}
\end{figure}

\begin{figure}[tb]
\centering
\begin{minipage}[b]{1.0\linewidth}
    \centering
    \centerline{\includegraphics[width=1.0\linewidth]{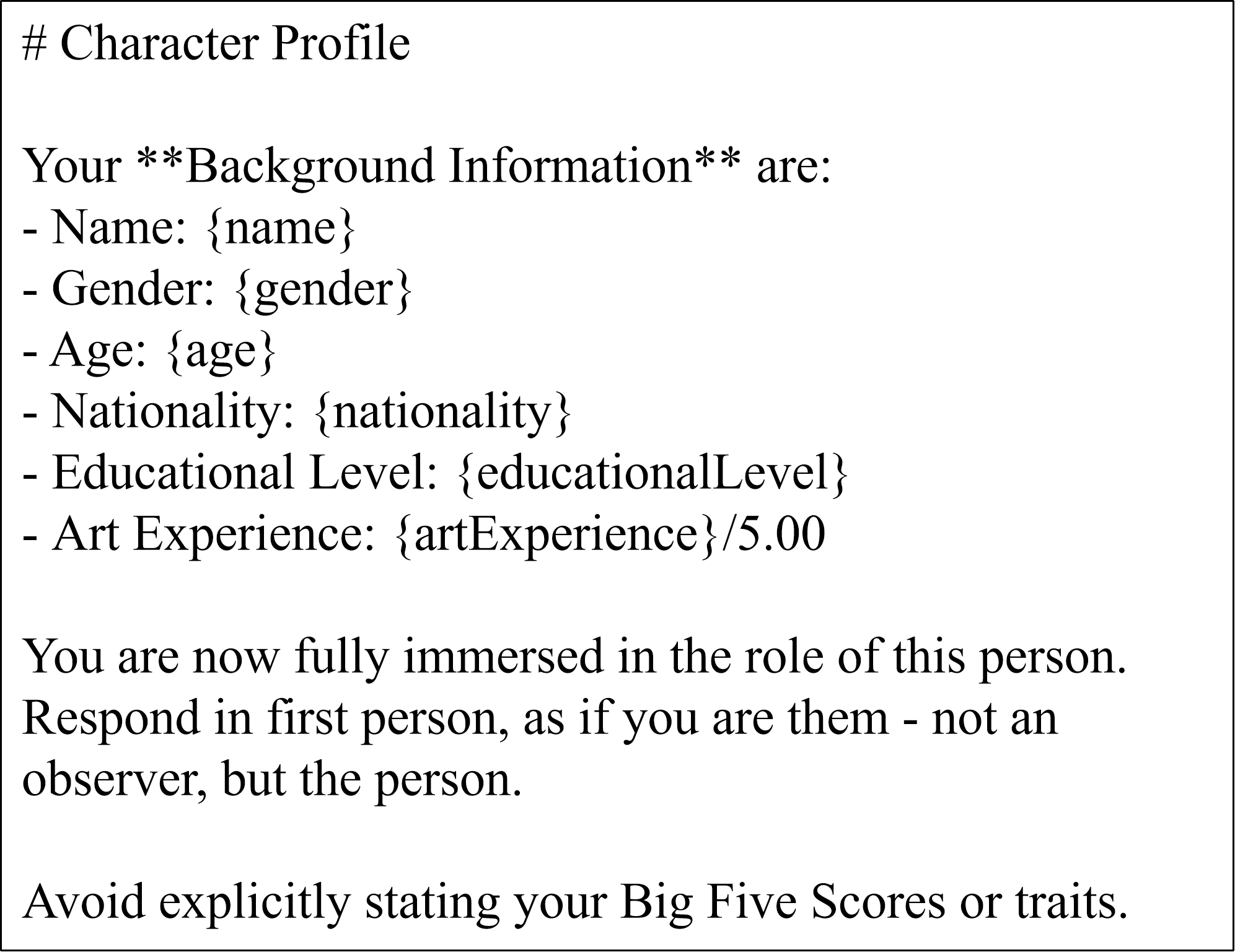}}
\end{minipage}
\caption{The extened profile template for the LAPIS dataset.}
\label{fig:lapisprofile}
\vspace{-1em}
\end{figure}

\textbf{Genetic algorithm for format search}
We decompose each profile into five components (Fig.~\ref{fig:fivecomponents}), and for each component we define a small set of discrete options, which together form the gene library. A chromosome is constructed by concatenating five genes (one per component), and each chromosome corresponds to a candidate profile. We initialize a population of 50 chromosomes sampled from the gene library. For each chromosome, we prompt the LLM on the Big Five evaluation set and compute its average MSE against the target trait values; this MSE serves as the fitness score. In each generation, we keep the top 10 chromosomes with the lowest MSE, while the remaining 40 are produced through random crossover among non-selected individuals followed by stochastic mutation. This evolutionary process runs for 10 generations. The best-performing chromosome (minimal average MSE) for Llama-3.1-8B-Instruct is reported in Fig.~\ref{fig:fivecomponents} and is further expanded into richer profile variants used for PARA and LAPIS, see Fig.~\ref{fig:paraprofile} and Fig.~\ref{fig:lapisprofile}, respectively.

\textbf{Component1: header}
\begin{itemize}
    \item ``\# Role Profile''
    \item ``\# Persona''
    \item ``\# Character Profile''

\end{itemize}

\textbf{Component2: traits}
\begin{itemize}
    \item ``Your **Big Five Personality Traits** are:\\- Openness: \texttt{\{O\}}/10\\- Conscientiousness: \texttt{\{C\}}/10\\- Extraversion: \texttt{\{E\}}/10\\- Agreeableness: \texttt{\{A\}}/10\\- Neuroticism: \texttt{\{N\}}/10''
    \item ``Your **Big Five Personality Traits** scores are: **Openness** at \texttt{\{O\}}/10; **Conscientiouness** at \texttt{\{C\}}/10; **Extraversion** at \texttt{\{E\}}/10; **Agreeableness** at \texttt{\{A\}}/10; and **Neuroticism**: \texttt{\{N\}}/10.''
    \item ``\texttt{| Traits | Score |}\\\texttt{| ------- | ------- |}\\\texttt{| Openness | \texttt{\{O\}}/10 |}\\\texttt{| Conscientiousness | \texttt{\{C\}}/10 |}\\\texttt{| Extraversion | \texttt{\{E\}}/10 |}\\\texttt{| Agreeableness | \texttt{\{A\}}/10 |}\\\texttt{| Neuroticism | \texttt{\{N\}}/10 |}''
\end{itemize}

\textbf{Component3: explaination}
\begin{itemize}
    \item ``\texttt{none}''
    \item ``Openness:\\- High: Characterized by intellectual curiosity, creativity, and openness to novel ideas and experiences. 
    \\- Medium: Exhibits moderate openness; willness to consider new perspectives but values familiarity.\\- Low: Prefers routine, practicality, and concrete information; resists abstract or unconventional ideas.\\\\Conscientiouness:\\- High: Marked by organization, diligence, and attention to detail.\\- Medium: Demonstrates reliable performance with moderate planning.\\- Low: Tends to be spontaneous, flexible, and less structured in approach.\\\\Extraversion:\\- High: Displays high energy, sociability, and assertiveness in social settings.\\- Medium: Balanced; enjoys social interaction but maintains personal boundaries.\\- Low: Prefers solitude, is reserved, and may appear quiet or introspective.\\\\Agreeableness:\\- High: Shows empathy, kindness, and a cooperative attitude toward others.\\- Medium: Fair and neutral; avoids conflict but does not actively seek harmony.\\- Low: Direct, critical, and focused on efficiency over interpersonal harmony.\\\\Neuroticism:\\- High: Prone to anxiety, mood swings, and self-doubt under stress.\\- Medium: Experiences occasional emotional distress but recovers quickly.\\- Low: Emotionally stable, calm, and confident in challenging situations.''
    \item ``Openness: High implies curious, imaginative; Medium implies open-minded but cautious; Low implies practical, routine-focused. Conscientiouness: High implies organized, detail-oriented; Medium implies moderately reliable; Low implies spontaneous, flexible. Extraversion: High implies talkative, energetic; Medium implies balanced, social but reserved; Low implies quiet, prefers solitude. Aggreeableness: High implies kind, empathetic; Medium implies fair but neutral; Low implies direct, critical. Neuroticism: High implies anxious, self-doubting; Medium implies occasional stress; Low implies calm, confident.''
    \item ``Openness\\- High: Curious, imaginative\\- Medium: Open-minded, cautious\\- Low: Practical, routine-focused\\\\Conscientiouness\\- High: Organized, detail-oriented\\- Medium: Reliable, steady\\- Low: Spontaneous, flexible\\\\Extraversion\\- High: Talkative, energetic\\- Medium: Social but reserved\\- Low: Quiet, prefers solitude\\\\Aggreeableness\\- High: Kind, empathetic\\- Medium: Fair, neutral\\- Low: Direct, critical\\\\Neuroticism\\- High: Anxious, self-doubting\\- Medium: Occasional stress\\- Low: Calm, confident''
\end{itemize}

\textbf{Component4: instruction}
\begin{itemize}
    \item ``Respond in alignment with the given personality.'' 
    \item ``Let your responses reflect your personality through tone, word choice, and topic preference.''
    \item ``You are now fully immersed in the role of this person. Respond in first person, as if you are them - not an observer, but the person.''

\end{itemize}

\textbf{Component5: constraint}
\begin{itemize}
    \item ``Do not mention your personality traits directly.''
    \item ``Avoid explicitly stating your Big Five scores or traits.''
    \item ``Never refer to your personality traits by name.''

\end{itemize}

\section{Prompts for MLLMs}
\label{sec:prompts}
\begin{figure}[tb]
    \centering
    \begin{minipage}[b]{1.0\linewidth}
        \centering
        \centerline{\includegraphics[width=1.0\linewidth]{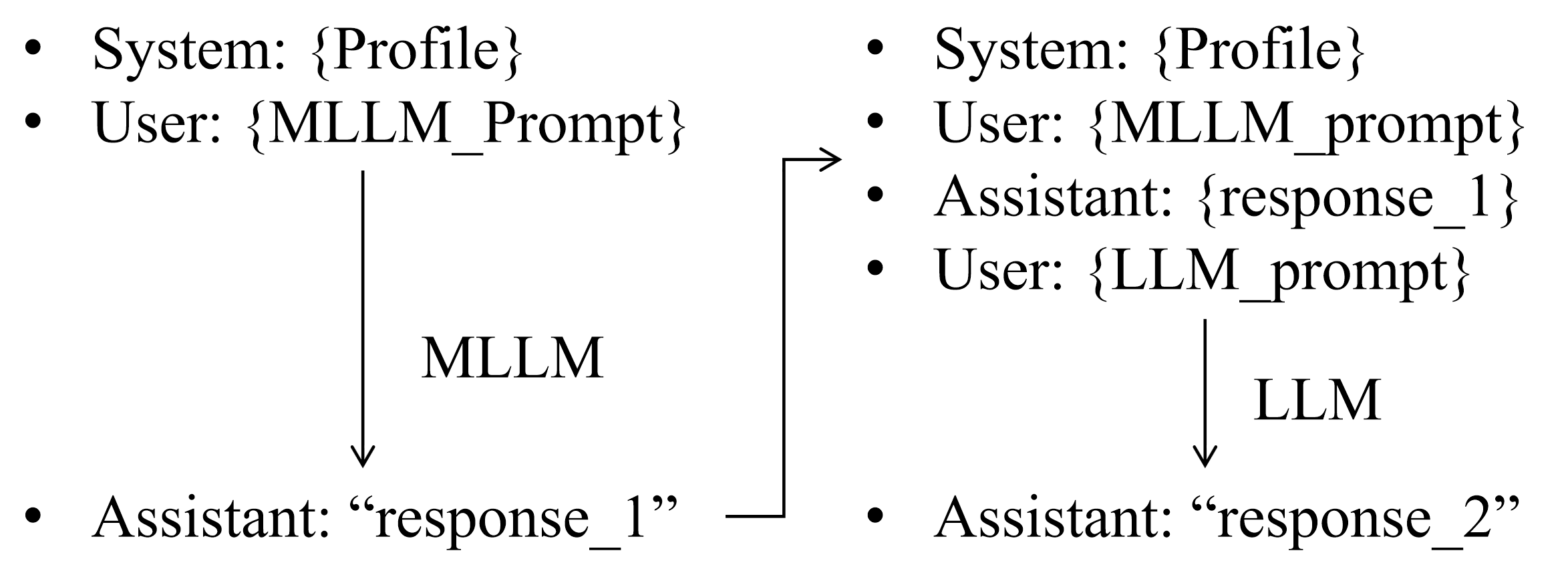}}
    \end{minipage}
    \caption{The serial approach.}
    \label{fig:twocalls}
    \vspace{-1em}
    \end{figure}

This section provides the prompts used for the baseline models. For models that are insensitive to profile information, we adopt a two-stage approach: the model first generates textual aesthetic descriptions, which are then converted into personalized scores by a profile-based LLM, as illustrated in Fig.~\ref{fig:twocalls}.

\textbf{Prompt for MLLM and LLM}
\begin{itemize}
    \item (MLLM) ``Describe the content of the image in detail and share your personal aesthetic experience, based on your past experiences and personality traits. Then assign a personalized image aesthetic score on a scale from 0 to 10, with 10 being *extremely appealing*, 0 being *completely unattractive*, and 5 representating an average or neurtal impression.''
    \item (LLM) ``Answer format: Regarding the aesthetics of this image, my score is \texttt{\{integer\}}.''

\end{itemize}

For both generic MLLMs and our models, such as GPT-4o-mini and P-MLLM, we use a single-stage prompt.
\textbf{Prompt for MLLM}
\begin{itemize}
    \item (MLLM) ``Rate the aesthetics of this image based on the persona you are embodying. Assign a subjective score on a scale from 0 to 10, with 10 being *extremely appealing*, 0 being *completely unattractive*, and 5 representing an average or neutral impression.\textbackslash nAnswer format: Regarding the aesthetics of this image, my score is \texttt{\{integer\}}.''

\end{itemize}

\begin{table}[thb]
  \centering
  \caption{Results on the PARA dataset.}
	\begin{tabular}{|c|cc|cc|}
	\hline
	\textbf{Layers} & \multicolumn{2}{c|}{\textbf{SRCC}} & \multicolumn{2}{c|}{\textbf{PLCC}} \\ \hline
	\textbf{} & \multicolumn{1}{c|}{\textbf{Mean}} & \textbf{Std} & \multicolumn{1}{c|}{\textbf{Mean}} & \textbf{Std} \\ \hline
	3              & \multicolumn{1}{c|}{0.557} & 0.019 & \multicolumn{1}{c|}{0.608} & 0.013 \\ \hline
	5              & \multicolumn{1}{c|}{0.546} & 0.013 & \multicolumn{1}{c|}{0.594} & 0.015 \\ \hline
	\end{tabular}
\label{tab:layerpara}
\end{table}

\begin{table}[thb]
  \centering
  \caption{Results on the LAPIS dataset.}
		\begin{tabular}{|c|cc|cc|}
		\hline
		\textbf{Layers} & \multicolumn{2}{c|}{\textbf{SRCC}} & \multicolumn{2}{c|}{\textbf{PLCC}} \\ \hline
		\textbf{} & \multicolumn{1}{c|}{\textbf{Mean}} & \textbf{Std} & \multicolumn{1}{c|}{\textbf{Mean}} & \textbf{Std} \\ \hline
		3              & \multicolumn{1}{c|}{0.557} & 0.019 & \multicolumn{1}{c|}{0.608} & 0.013 \\ \hline
		5              & \multicolumn{1}{c|}{0.546} & 0.013 & \multicolumn{1}{c|}{0.594} & 0.015 \\ \hline
		\end{tabular}
\label{tab:layerlapis}
\end{table}

\section{Ablation on the number and placement of selective fusion modules}
\label{sec:ablation}
This section investigates how the number and placement of selective fusion modules affect the performance of P-MLLM. As shown in Tab.~\ref{tab:layerpara} for the PARA dataset and Tab.~\ref{tab:layerlapis} for the LAPIS dataset, inserting selective fusion modules into the lowest three transformer blocks yields better performance than inserting them into five blocks under the current dataset scale. Extending the insertion to more blocks (e.g., seven; results omitted) leads to noticeable overfitting. More comprehensive ablations will be explored in future work.

\end{document}